# Order matters: Distributional properties of speech to young children bootstraps learning of semantic representations

**Philip A. Huebner (phueb001@ucr.edu)**
University of California Riverside
Interdepartmental Neuroscience Graduate Program , 900 University Ave, Riverside, CA 92521 USA

**Jon A. Willits (jon.willits@ucr.edu)**
University of California Riverside
Psychology Department, 900 University Ave, Riverside, CA 92521 USA

## Abstract
Some researchers claim that language acquisition is critically dependent on experiencing linguistic input in order of increasing complexity. We set out to test this hypothesis using a simple recurrent neural network (SRN) trained to predict word sequences in CHILDES, a 5-million-word corpus of speech directed to children. First, we demonstrated that age-ordered CHILDES exhibits a gradual increase in linguistic complexity. Next, we compared the performance of two groups of SRNs trained on CHILDES which had either been age-ordered or not. Specifically, we assessed learning of grammatical and semantic structure and showed that training on age-ordered input facilitates learning of semantic, but not of sequential structure. We found that this advantage is eliminated when the models were trained on input with utterance boundary information removed.

## Introduction

How do children learn complex linguistic and semantic knowledge? Constraints of some kind are necessary for narrowing the range of relations that need to be learned. Are these constraints a part of the learning and representational system, taking the form of innate linguistic knowledge (Pinker, 2003), a specialized language learning ability (Gleitman, 1984; Newport, 1990), or domain-general learning mechanisms (Elman et al., 1996; Gentner, 1983; Saffran et al., 1997)? Alternatively, constraints may be a part of the input itself. For example, highly structured linguistic experiences can result from constraints on language production and communicative factors (MacDonald, 2013).

One notable constraint on the input is the linguistic simplification provided by child-directed speech (CDS) (Gleitman, et al., 1984), which is characterized by larger pitch contours and lengthened vowels (Fernald & Kuhl, 1987), and a restricted range of conversational topics, simplified sentence structures, and longer pauses between utterance boundaries (Gallaway & Richards, 1994). In these and other ways, CDS is thought to facilitate language development by presenting an initially restricted hypothesis space, and then gradually expanding it (Cameron-Faulkner et al., 2003). While numerous benefits of CDS have been found on early language acquisition (Golinkoff and Alioto, 1995), others have shown that children appear to learn language just as well when their primary caregivers do not use CDS (Schieffelin and Ochs, 1986). It is unknown how beneficial CDS is for learning linguistic structure and word meanings from language, and if it is beneficial, why?

Computational modeling has been used to try to resolve this debate, but results are inconclusive. Elman (1993) showed that a simple recurrent neural network learns the structure of a hierarchically-organized artificial grammar better when either the input or the model is constrained in such a way that simpler grammatical relationships are learned first. However, Rohde & Plaut (1999) showed across a wide range of artificial training corpora that "starting small" tends to hinder learning more frequently than it provides a benefit. Since the time of Rohde and Plaut's study, the issue has remained largely untouched.

Studies investigating whether CDS can facilitate lexical semantic development are largely absent in computational modeling studies. A number of models have been shown to be quite good at learning semantic structure from speech or text input (see Jones, Willits, & Dennis, 2016 for review). However, the majority of these models are insensitive to the age-related asymmetries inherent in CDS. Any benefit of a gradual increase in linguistic complexity or semantic content can only be observed in models sensitive to the order of linguistic input, and the most widely tested distributional models are not. The simple recurrent network (SRN) on the other hand, is naturally sensitive to training order, and previous research has shown that it can learn linguistic and semantic structure from CDS input (Huebner & Willits, 2018).

Using the SRN, we revisited the "starting small" hypothesis in the context of learning both the sequential and semantic structure of CDS. We used the CHILDES corpus (MacWhinney, 2000) as input, which, as we demonstrate, exhibits "starting small" properties typically associated with CDS. We hypothesized that taking advantage of the age-ordering of CHILDES may improve learning of grammatical and/or semantic structure. This work builds on Elman's and Rohde and Plaut's work using artificial grammar. However, given that our corpus is a collection of naturalistic CDS transcripts ordered by the age of the speech recipient, the input "starts small" out-of-the-box, which avoids questions about which input is "most natural".

The aim of the paper is twofold. In Study 1 we demonstrate that the CHILDES corpus does indeed "start small", and how. In Study 2, we investigate whether a model trained to predict sequential structure can benefit from training first with speech to younger children. Does the model learn to predict word sequences better? Are the learned internal representations more useful for semantic classification?

## Study 1

Study 1 is a corpus analysis of child-directed speech in the CHILDES corpus designed to quantify the extent to which it "starts small".

## Methods

### Corpus Preparation

As a representative sample of naturalistic speech to children, we used the CHILDES corpus, transcripts of interactions with children in various situations (MacWhinney, 2000). We used all transcripts involving typically-developing children 0 to 6 years of age from American English speaking households, and excluded those for which no age information was available. This resulted in 3251 transcripts containing 22,448 types, and 5,113,856 tokens. Our version of the corpus was obtained from childes-db.stanford.edu on Dec 1, 2017.

The corpus was tokenized into individual words using the python module *spacy*, which split the corpus on spaces and contractions, and was converted to lowercase. We left sentence-boundary punctuation (periods, question marks, etc.) in the corpus in one set of simulations, and removed them entirely in another. We performed no further processing, to leave intact as many naturalistic properties of the corpus as possible.

We ordered the transcripts in the corpus by the age of the target child, concatenated all transcripts into a single composite corpus, and then split this corpus into 256 equally sized partitions (containing 19,976 words each). This number was chosen so that the number of words in each partition roughly corresponding to the amount of words an English learning child hears in a day (Hart & Risley, 1995). Two different corpora where then created from these partitions, a chronological corpus that maintained the chronological age of the target child, and a shuffled corpus in which the order of the partitions was randomized, and thereby removing any age based asymmetries in the corpus.

## Results

### Age-related Asymmetries in CHILDES

There are many age-related differences in the CHILDES corpus. Here, we discuss those most related to the claim that speech in younger children is structurally less complex.

First is the frequency of novel n-grams. We calculated all n-grams in the range 1 to 6, tracking the number of novel n-grams encountered proceeding from start to finish through the chronologically or randomly ordered corpus. THe motivation behind this analysis was to test whether speech to younger children contains high frequency token patterns that are more frequently repeated compared to speech to older children. If this is the case, then the number of novel n-grams should be lower in the early part of the chronological corpus compared to the random-order corpus.

The top panel of Figure 1 plots the number of unique words encountered over the course of both corpora. The section containing the first 10,000 words is enlarged at the top right. This panel shows that, at the earliest stage of the corpus, occurrences of new words are less frequent in the chronological condition (in blue) compared to the shuffled condition (in green). For single words, this effect is entirely driven by the very earliest stage of the input (i.e. speech to children under 1 year of age). The size of this difference grows both in magnitude and duration, as the size of the n-gram grows. For 6-grams, children are encountering dramatically fewer unique 6 word sequences in early speech, and this difference persists almost halfway through the corpus (the blue curve crosses over the green curve approximately ⅓ of the way through the corpus, where it aligns with speech to 2-year-olds). Thus, as late as 2 years of age, speech to children is still simpler and less variable, in terms of 6-grams, than would be expected due to chance.

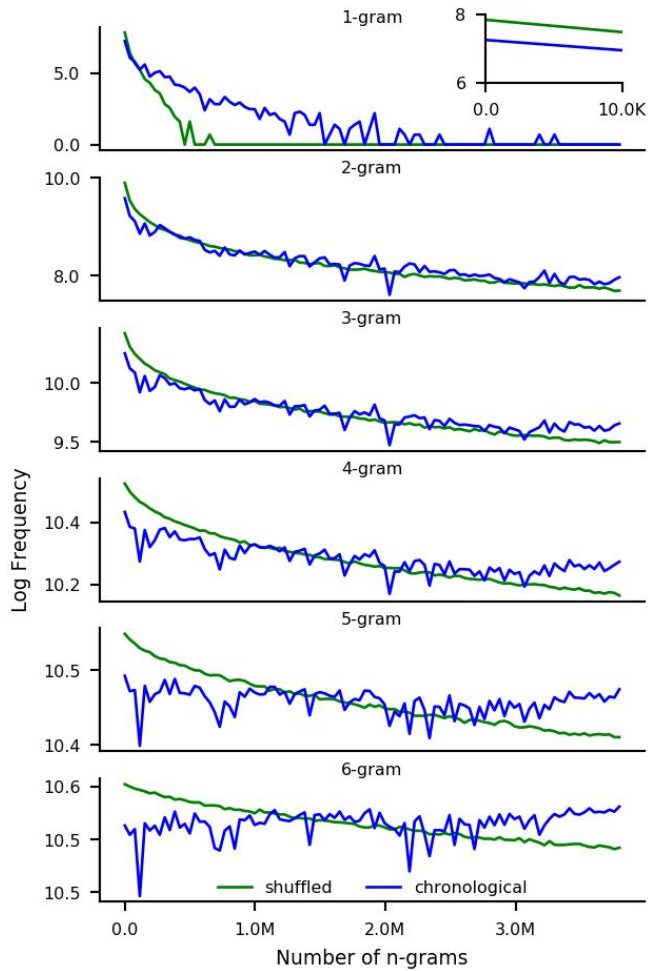

**Figure. 1.** Frequency of novel n-grams as function of corpus location. Direction of analysis is left to right.

Next, we investigated measures of age-related asymmetries in utterance complexity. We computed the

Shannon entropy for each corpus partition. Similar to the type-token ratio, a low value indicates frequent usage of high-frequency compared to low frequency types. We also analyzed the corpus using a rolling window analysis of the mean and standard deviation of utterance length across the corpus. These analyses (Figure 2) demonstrate that all three measures gradually increase in CHILDES.

Last, we used the Python library *spacy* to tag all tokens with their grammatical category, producing a list of the nouns, verbs, and adjectives in the corpus. We then computed the average corpus location of occurrence for each word. For example, *bottle* is a noun spoken much more frequently to younger children, with a low average location, and *story* is a word spoken much more frequently to older children, with a higher average location. We took each list of words within each grammatical category, sorted the words by their average location, and split these lists into two halves (1st half dominant vs. 2nd half dominant). We then counted the total token frequencies of the words in each split at binned intervals in the corpus.

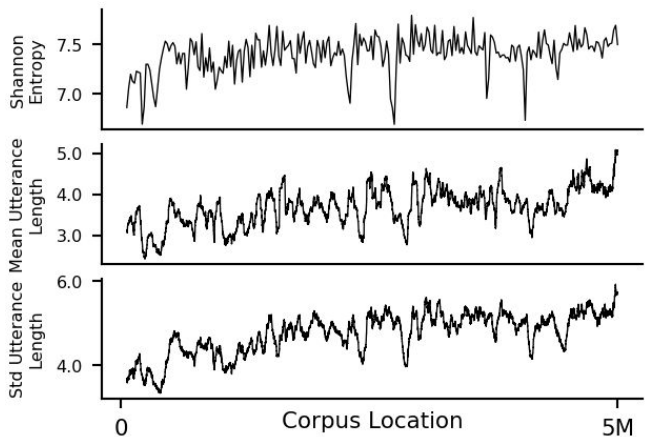

**Figure. 2.** Rolling Window Corpus Analysis. A) Shannon Entropy. B) Mean and Standard Deviation of Utterance Length

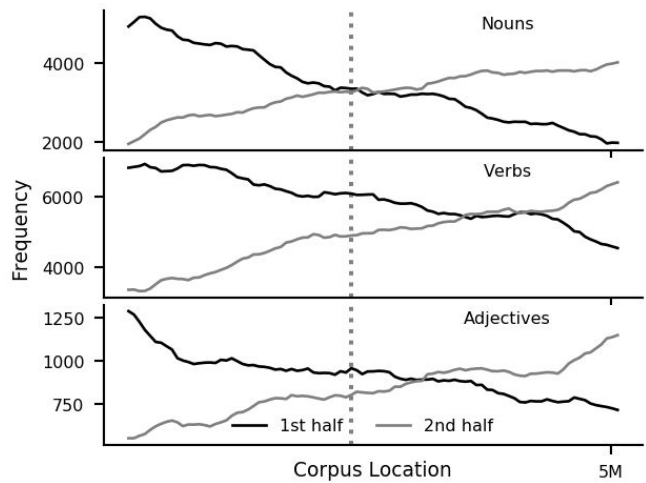

**Figure 3**. Word Frequency as function of corpus location separated by Grammatical Category (panels) and mean corpus location (black vs. grey line).

These counts are plotted in Figure 3, revealing a number of interesting age-based asymmetries. First, all 1st-half biased words are much more frequent than 2nd-half biased words in the early portion of the corpus. For example, the area under the curve representing the total frequency of 1st-half biased nouns is larger than the area under the curve representing 2nd-half biased nouns, which shows that the early input is more noun-dominant. This indicates that more words belonging to grammatical categories are introduced in the later part of the corpus.

Also, examining where the two curves cross in each panel relative to the corpus midpoint (indicated by the vertical dotted line), reveals that both 1st-half biased verbs and adjectives tend to be reused across longer distances in the corpus. This means that words occurring early in the chronologically ordered corpus are not only repeated more often, but are also more spread out across the corpus compared to words that occur relatively late.

The above analyses clearly demonstrate that the CHILDES corpus exhibits "starting small" characteristics when ordered by the age of the target child. The speech to young children in CHILDES has fewer novel n-grams, shorter utterance length and variance, lower lexical diversity, and less frequent occurrences of grammatical words.

## Study 2

Having established that CHILDES has age-related asymmetries in lexical and grammatical complexity, we test whether these asymmetries facilitate learning of sequential structure, and semantic representations in a recurrent neural network.

## Methods
### Model Architecture
Our model is based on the simple recurrent network (SRN), containing an input and output layer (with one unit in each layer for each word), and a hidden layer between the input and output layer, encoding the model's internal representations for each word. The model has weighted connections from the input layer to the hidden layer, from the hidden layer to the output layer, and from the hidden layer back to itself (Elman, 1990). The recurrent connectivity allows the SRN to integrate information from previous time steps, and enables learning of sequential statistics in the input. In order to predict complex structures and combinatorial dependencies, the model must learn weights that lead to useful patterns of activation in all of its layers.

The SRN has 4096 input and output units, one for each of the 4095 most frequent words in the corpus, and one out-of-vocabulary unit for all of the remaining words (approximately 1.9% of all tokens). This resulted in each vocabulary item occurring at least 10 times during training. Words were presented to the model one at a time, in order. A presented word activates its input layer unit, and this activation is propagated forward to the hidden layer (containing 512 units) via weighted connections. Each hidden layer unit also receives weighted input from the recurrent layer at the previous time step, and the sum of the current and recurrent input is transformed by the nonlinear sigmoid function, resulting in the hidden layer's activation. Finally, each hidden unit propagates its activation to the 4096-unit output layer via weighted connections. The pattern of activations at the output layer was transformed using the softmax operation, giving a probability distribution representing the SRN's prediction of the next word, given the current input.

**Training Procedure**
We trained the SRN using truncated backpropagation through time (Werbos, 1990) and 20 iterations over each corpus partition. First, an input sequence consisting of 7 consecutive words was fed to the model as explained above, and output layer activations were computed. Next, these activations were compared to the actual next word in the corpus, and this information is used to create an error signal to adjust the weights in the network, so that on future occurrences it is more likely to predict the observed output given the observed input. Through this procedure, the network starts with a set of initially randomized weights (resulting in a flat output probability distribution at the beginning of training), and converges on weights that will be more effective at predicting sequences, given that there is statistical structure in the input. This process repeats until all 7 word windows have been presented to the model.

We trained the model using the open-source machine-learning framework TensorFlow (Abadi et al., 2016), and the code, including the corpus are available at https://github.com/phueb/rnnlab.

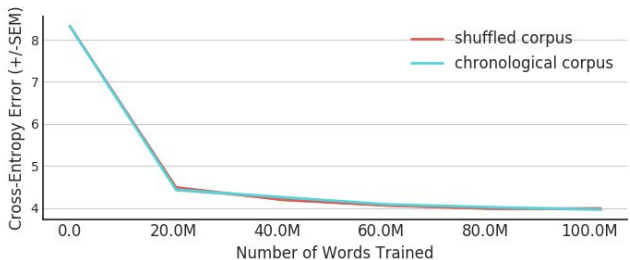

**Figure. 4**. Trajectory of cross-entropy error (which is measured on a natural log scale) as a function of training order. The two trajectories almost completely overlap.

## Results
### Sequential structure prediction
To evaluate sequential structure prediction performance as a function of training on either the chronological or shuffled corpus, we evaluated the SRNs on the error measure used during training, its cross-entropy. A higher value cross entropy indicates larger prediction errors. We evaluated each model before, 4 times during, and once after training. At each evaluation point, each model was evaluated on its predictions for each word in the corpus, given its context window. An untrained model predicting each word in the corpus receives a score roughly $ln(vocabulary\ size) = 8.3$.

We trained five SRNs (starting from different randomly initialized weights) on the chronological corpus, and five on the shuffled corpus. At the end of training, the average cross-entropy score of all SRNs was $4.0 \pm 0.2$ (M ± SD) in the chronological training condition and $4.0 \pm 1.2$ (M ± SD) in the shuffled training condition (Figure 4). It is clear that learning of word sequences has taken place in both models; however, no statistical difference between the two conditions was observed. Thus, starting with speech to younger children from the CHILDES corpus is not better for learning word sequences in the SRN.

### Semantic Category Knowledge
To test the model's acquisition of semantic knowledge, we evaluated the most frequently occurring nouns in the vocabulary (excluding any that occurred less than 10 times in the corpus) and that were judged to belong unambiguously to a set of 29 semantic categories (such as *mammal*, *vegetable*, *clothing*, etc.) according to 20 human raters. This resulted in a collection of 563 probe words, for which we calculated internal representations.

These semantic representations were calculated using a 2x3 design. The first manipulation was the corpus training condition (chronological or shuffled) described above. The second manipulation involved three ways of deriving representations from the models. The first was "ordered context" in which the model was provided with all 7-word windows containing a probe in the final position to produce a hidden state activation for that probe in its context. These hidden states were then averaged together to obtain a single representation for a given probe. The second was "shuffled context", calculated the same way as above, but with context words presented to the model in a randomized order. The third evaluation condition we termed "without context", and in this condition the model was provided with the probe word only. We investigated all three evaluation conditions, as each provides insight to a different representational ability: The first is a measure of how well the SRN can derive meaning from ordered contexts; the second shows how the model can interpret meaning when contexts do not contain word order information, and the third is a measure of the model's ability to learn de-contextualized, abstract word representations, and a test of much the model's semantic representations depend on the context in which the probe words occur.

We evaluated these six model conditions (two corpus training conditions and three semantic evaluation type conditions) across six time points (one pre-training, one post-training, and four at equally spaced points during training), creating a 2x3x6 mixed design, with training corpus as a "between models" comparison, and semantic evaluation type and time point as "within model" comparisons. As in the previous analyses, five models were trained for each "between model" comparison.

The dependent measure in the analysis was accuracy at classifying two words as belonging to the same category, based on the similarity of their representations (Huebner & Willits, 2018). Judgements are based on the 563x563 probe similarity matrix $S$ obtained by computing all pairwise similarities between probe representations (calculated separately for the "with ordered context", "with shuffled context", and "without context" conditions). We constructed matrix $S$ at the same time points during training at which we evaluated cross-entropy to be able to observe any learning differences during training. Each probe-probe similarity in matrix $S$ was used to make a "same vs. different" judgment within a signal detection framework, tested at multiple similarity thresholds ($r$ = 0.0 to 1.0 with step size 0.001) to determine the threshold for maximum accuracy. In other words, if two probe words $i$ and $j$ belong to the same category, and if $S_{i,j} > r$, a "hit" is recorded, whereas if $S_{i,j} < r$, a "miss" is recorded. On the other hand, if the two probe words do not belong to the same category, either a "correct rejection" or "false alarm" is recorded, depending on whether $S_{i,j} < r$ or $S_{i,j} >$ r. After calculating the sensitivity and specificity for each comparison, we averaged the two to produce the balanced accuracy for each probe, eliminating bias resulting from the fact that a vast majority of probe pairs do not belong to the same category. The measure of interest is the average of all the probes' balanced accuracies at the similarity threshold which yielded the highest value.

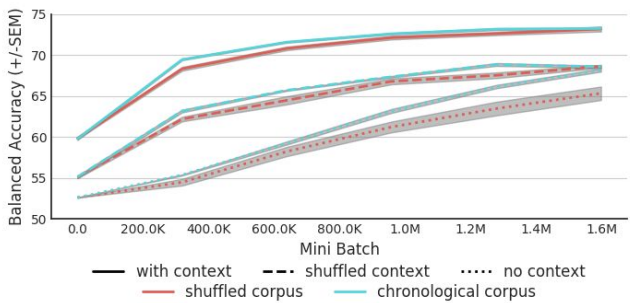

**Figure. 5**. Trajectory of semantic classification task performance, measured using balanced accuracy, with error bars (std) in grey.

We used this process to compute a balanced accuracy score for each model, in each evaluation condition, at each time point. These scores are shown in Figure 5. Statistical analysis revealed a three-way interaction ($F(10,40) = 4.89$, $p < 0.001$). Follow-up analyses showed this interaction to reflect the following: First, performance in all conditions gradually increased with more training, but at different rates in each evaluation condition. The best overall performance was achieved when semantic classification accuracy was evaluated in the "ordered context" condition, ending at 73.2% ± 0.14% (M ± SD). While we found no difference at the endpoint between the chronological and shuffled corpora in this condition, the chronological condition showed a slightly faster growth. The "shuffled context" evaluation condition performed second best, ending at 68.2% ± 0.14% (M ± SD), again with no difference at the endpoint between the chronological and shuffled corpora, and slightly faster growth when the chronological corpus is used. The "no context" evaluation condition performed the worst overall. Here, we did observe a large difference between the chronological and shuffled corpora. The models trained on the chronological corpus performed much better than the models trained on the shuffled corpus, with both a faster growth trajectory and a higher endpoint (68.1% ± 0.59%, versus 65.2% ± 1.01%).

The benefit that contextual information provides when retrieving semantic information about a word is consistent with previous literature, such as the word superiority effect (Paap et al., 1982). It makes sense that the semantic representations activated by the model would be better with context, as this can provide supporting semantic information. The lack of contextual information is arguably most representative of learning an "abstract" representation of a word's meaning, and it is notable that this is where "starting small" appears to show the largest benefit.

**Removal of Utterance Boundary Markers**
Of final interest is the effect of utterance boundary markers (periods, question marks, etc.) on model performance. While they are obviously not present in child-directed speech, many acoustic features are highly correlated with utterance boundaries. Because the mean utterance length gradually increases, it is necessarily true that punctuation tokens are over represented in the earliest portion of the corpus. Indeed, the utterance boundary marker '.' is the most asymmetrically (right-skewed) distributed of all vocabulary items. This suggests that punctuation might play a special role in organizing early representations in the chronological training condition. To investigate this, we trained a new set of models on the same corpus but with utterance-boundary markers removed. The results of this analysis are shown in Figure 6.

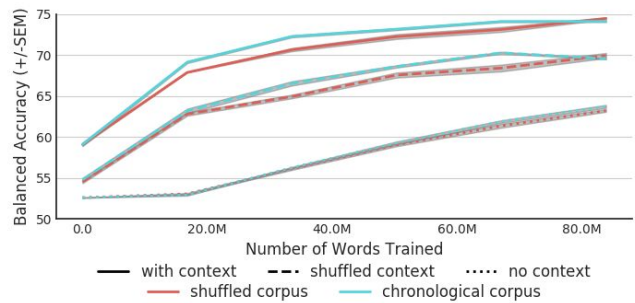

**Figure 6**. Trajectory of semantic classification performance, for models trained with utterance boundary punctuation removed.

Again, we found a three-way interaction ($F(10,40) = 8.72$, $p < 0.001$). Critically, the benefit of training in chronological order in the "no context" evaluation condition, was eliminated. The learning benefit provided by punctuation is consistent with findings by Mintz et al. (2002) who showed that word contexts can be more informative for category learning when they do not cross phrase boundaries. It is likely that the model can use utterance boundary information to constrain what part of a word's context might be most useful for learning about its semantic category.

## Discussion

The aim of this paper was twofold. First, we wanted to show that the CHILDES corpus "starts small". We showed this is the case for utterance length and variance and that these measures gradually increase over the course of the corpus. We also showed that the early portion of CHILDES contains fewer n-grams, indicating reduced structural complexity, and the same words are repeated more often compared to later portions of the corpus.

Our second goal was to assess whether the simplification of linguistic input provided by CDS can produce an observable benefit for computational models of language and semantic development. Rohde & Plaut (1999) found that "starting small" is most effective when important linguistic dependencies span uninformative clauses in an artificial language corpus, arguing therefore that "starting small" should not benefit learning of naturalistic language. Indeed, we found no evidence that training over age-ordered CDS gradually increasing in complexity improves learning of sequential structure. However, we did find a beneficial effect on a semantic classification task. This effect was only observed when semantic representations were accessed without context. We also showed that removal of punctuation eliminates this effect. Training with utterance boundary cues likely helps the model to constrain what contextual information is most useful for learning.

Three important questions remain. First, why does the order of linguistic experiences matter for learning of semantics but not syntax? It is possible that our measure of syntactic learning, based on the cross-entropy error, is not psychologically appropriate given that it measures the model's fit to the data rather than some deeper structural properties of language. Furthermore, since cross-entropy is an average measure over all words in the corpus, it simply might not be sensitive enough.

Second, why did we observe an advantage for the acquisition of abstract, (non-contextualized) and not contextualized representations? One possible answer is that abstraction requires learning of higher order features of the input, and therefore "starting small" might direct the model's representational trajectory towards learning those higher order features earlier (Clark, 1994).

Third, what are the specific reasons that training on the chronologically ordered corpus yielded better results? We have argued that the effects are due to "starting small", and specifically due to the organizational role of early boundary cues. More work is needed to narrow down the exact mechanism, and to improve our understanding of the benefits, and limitations that "starting small" might provide to both computational models and infants.